# A Possible Artificial Intelligence Ecosystem Avatar: the Moorea case (IDEA)

Jean-Pierre Barriot[1], Neil Davies[2], Benoît Stoll[1], Sébastien Chabrier[1] and Alban Gabillon[1]

[1]Géopôle du Pacifique Sud, University of French Polynesia (GePaSUD) – Outumaoro Campus, BP 6570, 98702 Faa'a, Tahiti, French Polynesia, email: jean-pierre.barriot@upf.pf
[2]Gump Research Station, University of California, Berkeley (GUMP Station) – Maharepa, Moorea Island, BP 244 - 98728, French Polynesia

**Abstract**

High-throughput data collection techniques and largescale (cloud) computing are transforming our understanding of ecosystems at all scales by allowing the integration of multimodal data such as physics, chemistry, biology, ecology, fishing, economics and other social sciences in a common computational framework. We focus in this paper on a large scale data assimilation and prediction backbone based on Deep Stacking Networks (DSN) in the frame of the IDEA (Island Digital Ecosystem Avatars) project (Moorea Island), based on the subdivision of the island in watersheds and lagoon units. We also describe several kinds of raw data that can train and constrain such an ecosystem avatar model, as well as second level data such as ecological or physical indexes / indicators.

1. **Introduction: what is an ecological avatar?**

Unlike some aspects of climate change, processes related to biodiversity and ecosystem services are typically place-based. Inspired by successes in modeling complex systems at other scales of organization, notably the cell (Karr et al., 2012), the Island Digital Ecosystem Avatars (IDEA) Consortium aims to build computer simulations ("avatars") to the scale of whole social-ecological systems. With a common boundary constraining their physical, ecological, and social networks, islands have long been recognized as model systems for ecology and evolution. This is why the island of Moorea (French Polynesia) was selected to build such an "avatar". The Moorea IDEA aims to understand how biodiversity, ecosystem services, and society will co-evolve over the next several decades on this island, depending upon what actions are taken. Specifically: (1) what is the physical biological, and social state of the island system today? (2) How did it get to this point? (3) What is its future under alternative scenarios of environmental change and human activity, including conservation efforts? The IDEA team then defined the Moorea Avatar as the best digital representation of the island; a three-dimensional visualization of Moorea that will look similar to the one on Google-Earth, but which will include the dimension of time and enables researchers to zoom into a location, access data, and run simulations (Davies et al., 2016). In this paper, we will mainly focus on the point (3) of this program, with implications about points (1) and (2). This means that we will restrict the "avatar" as a predictive model that relies on two points: (a) the best possible description of the current state of the island ecosystem, (b) the belief that the future state of the ecosystem can be derived, on a stable manner (as opposed to "chaotic" in the mathematical sense) from the present state. In paragraph 2, we will present an historical ecological model built from differential equations, which is already highly non-linear. In paragraph 3 we present Artificial Neural Networks as an Artificial Intelligence extension of differential equations, and in paragraph 4 we present our proposed application to the Moore IDEA case Avatar.

2. **Ecosystem models**

Ecosystem modelling takes its root a century ago, with the Lotka–Volterra predator–prey model (Volterra, 1926, Berryman, 1992), a pair of coupled highly non-linear differential equations with periodic solutions, in the form (1):

$$\frac{dr}{dt} = r(\alpha - \beta f) = \alpha r - \beta r f$$

$$\frac{df}{dt} = f(-\gamma + \delta r) = \delta r f - \gamma f$$

where *r(t)* and *f(t)* are the prey (like rabbits) and predator (like foxes) populations over time t, $\alpha$ is the prey-population grow rate in the absence of predators, *β* is the predator attack rate, *γ* is the death rate of predators in the absence of their prey resource, and *δ* is the prey-induced birth rate on predators. All these $\alpha, \beta, \gamma, \delta$ parameters, to be solved, are by essence positive and describe the interaction of the two populations on a closed ecosystem. This model was highly successful and was applied to many areas of ecology and even economics (May and McLean, 2007). Its limitations are obvious: in the absence of predators, the growth of the population prey is exponential (i.e. unlimited food supply), and predators of predators are ignored (like fox hunting).

We can already take lessons from this model: from a set of observations of the two populations (r(t₁), r(t₂),….) and (f(t₁), f(t₂)…) let us extract the parameters α, β, γ, δ. This is basically a linear process, as we can write the system (1) as (2):

$$\begin{bmatrix} r & -rf & 0 & 0 \\ 0 & 0 & rf & -f \end{bmatrix} \begin{bmatrix} \alpha \\ \beta \\ \gamma \\ \delta \end{bmatrix} = \begin{bmatrix} dr/dt \\ df/dt \end{bmatrix} \quad ==> \quad Gp = d \text{ for any given data point.}$$

Here we approximate the derivative $\frac{dr}{dt}$ by the crude approximation $\frac{dr}{dt} = (r(t + \Delta t) - r(t))/\Delta t$ for a given time step Δt (same for $f$). If we do not inforce the positivity of the parameters α, β, γ, δ, the full array of L₂ norm optimization (i.e. "fitting") can come into play to obtain an estimate of $p = (\alpha, \beta, \gamma, \delta)$ with a cost function of the form $Q(p) = \|Gp - d\|^2$ + penalty terms, if any. In a modern terminology, it is an *unsupervised learning* process, where the time series (r(t₁), r(t₂),….) and (f(t₁), f(t₂)…) of the observed populations permit to build an estimate of the parameters α, β, γ, δ which is as reliable as the length and accuracy of the time series. Once we have a "reliable" estimate of the parameters α, β, γ, δ, we can use Eq. (1) to predict the evolution of the populations of prey and predators, up to a "reasonable" time horizon.

We can recast this "fitting" process as finding an operator M linking, in the best possible way, the two time series $U = (..., (r(t_i), f(t_i)), ...)$ and time series $V = (..., (r(t_{i+1}), f(t_{i+1})), ...)$, in the form M(U) = V. This is called a *supervised learning process*. In the L₂ norm this means minimizing $Q(M) = \|M(U) - V\|^2$ w.r.t. to the operator M. Of course, if we write M(U) = M_{αβγδ}(U) = V, i.e. if we suppose that M as the underlying structure of Eq. (2) and minimize Q(M_{αβγδ}) = Q(α, β, γ, δ) w.r.t. α, β, γ, δ, we are then back to the previous solution. I we do not now suppose *anything* about M, the solution becomes severely ill-posed, as shown by practice, because Q presents many local minima w.r.t M. The minimum is nevertheless unique is we suppose that the operator M is linear, i.e. that V= M(U) = M U in the matrix sense, resulting from the equality (3):

$$Q(M) = Trace((MU - V)(MU - V)^T) = Trace((VU^+ - M)UU^T(VU^T - M)^T + V(I - U^TU)V^T),$$

Once built, the operator M can be used as a predictor, in the form of (4):

$$y = M(x) = V U^+ x,$$

where U⁺ is the Moore-Penrose inverse of U (Campbell and Meyer, 1979, Douglas and Miranker, 1990). The vectors $x$ and $y$ represent the present and future states of the system. This result can be recast in a probabilistic framework (Barriot et al., 1999), by seeing U and V as samples of an underlying stochastic process. For example, the inputs (respectively outputs) can be the number of rabbits (respectively foxes) in the 58 counties of California for the last 50 years. In that case, U and V are matrices with 58 lines and 50 columns, and M is a 58 x 58 square matrix with rank 50 (i.e. not invertible). But even if M is of full rank (i.e. with more years of observations than counties), M cannot model the relation between prey and predators, because the solution of Eq. 1 is periodic, so we can have the same number of foxes for two different numbers of rabbits for each county. Even the "naïve" prey-predator model is already highly non-linear with respect to r and f (but it is linear w.r.t. *α, β, γ, δ)*. It should be also stressed that even periodical forcings (like the seasonal availability of food) does not imply periodical solutions (Floquet's theorem on the stability of differential equations, see for example Barriot (2015)). The pseudo-inverse of U is very often numerically unstable, and it is preferable to use a regularized form (i.e. numerically stable and carrying the same information contents), by using a truncated singular decomposition or a Tikhonov regularization (Tikhonov, 1977) with a non-negative ridge parameter, according to (5):

$$U^+_{[\lambda]} = [U^T U + \lambda I]^{-1} U^T$$

Highly non-linear systems of coupled differential equations are known for centuries in physics, like the Navier-Stokes equations for weather modeling (McGuffie, 2005, pp 167-169), with many "free" parameters (e.g. the specific heat of air), but their exact mathematical form is derived from extremely well seated laws of physics. Ecosystem models are by their whole nature a lot more "fuzzy". A well-know, actual world, semi-empirical ecosystem model is Ecopath (Bevilacqua et al., 2016), which creates a static mass-balanced snapshot of an oceanic ecosystem represented by trophically linked biomass "pools". Ecopath data requirements are relatively simple: biomass estimates, total mortality estimates, consumption estimates, diet compositions, and fishery catches. Let us note that the differential equations used on these empirical or semi-empirical models can be generalized by cellular automata approaches where sets of *a priori* rules (including free parameters) operate on discrete spatial grids (Hogeweg, 1988).Recently, differential equations models experienced a new start, with the introduction of resilience patterns (Gao et al, 2016).

3. **Neural Networks: An approach to Artificial Intelligence**

When the fuzziness becomes large, which is often (but not always) synonymous with "large" problems, scientists are often left with a string of question marks about the exact relationships between the different "players" showing up in the ecosystem. Artificial Neural Networks (ANN) emerged for this type of "fuzzy" applications in the late fifties (Rosenblatt, 1958). Artificial neural networks are generally presented as systems of interconnected "neurons", mimicking in some way the brain. The basic inner working of a single (column-wise) layer of neurons can be mathematically defined as follows (6):

$$y = \sigma(Wx)$$

where $x$ and $y$ have the same meaning as in Eq. 4 and $W$ is a matrix of weights acting linearly on $x$, $\sigma$ is an activation function acting on $Wx$. Examples of activation function are the step function (0 for $R^-$, 1 for $R^+$) or sigmoid functions (smoothed, continuous versions of the step function), or the now popular rectified linear unit (ReLU) of Teh and Hinton (2001) with $\sigma(x) = \max(0, x)$. The activation function mimics brain activity. For example "firing" of the neuron ($\sigma = 1$) occurs for the $\sigma$ step function if its argument becomes larger than 0. ANNs are formed by stacking layers of neurons, the outputs of one layer forming the inputs of the succeeding layer (each neuron output of the preceding layer being duplicated for part or all the neurons of the succeeding layer), with of course the exception of the first layer (ANN input-at-large) and the last layer (ANN output-at-large). If we drop the activation function (i.e. $\sigma(x) = x$) we go back to Eq. (4) for the determination of the weights $W$ with, in this case, $W = M$. The difficulty with Eq. (6) lays, of course, in the determination of $W$ when the activation function is present, which is a mandatory requirement to model highly non-linear systems like the prey-predator system of Eq. (1). The usual way to determine $W$ in ANNs is through a mechanism known as "backpropagation" (Alpaydın, 2010), a gradient descent from a priori $W°$ to an "optimal" $W$ by following the gradient $\nabla_W Q$. This approach is supposed to find an acceptable minimum to the cost function $Q(W)$, but mathematically, this approach only guarantees to find a local minimum, close to $W^0$ (convexity problem). There is no known algorithm to find the minimum minimorum of Q and the question to determine good "starting" a priori $W^0$ stays open. After some initial successes with shallow ANNs (i.e. ANNs with a few layers) during the eighties, this research field became stagnant (the AI "winter") because of these two unsolved points and also a lack of computing power (Bengio et al., 1994, Bengio, 2009 and 2016).

The field of ANN research was revived in 2006, when the optimization difficulty associated with deep (as opposed to shallow) ANNs was empirically alleviated by the efficient learning scheme introduced by Hinton et al. (2006a, 2006b). Today, the field of research around Deep Neural Networks (DNN) is in a boiling state (LeCun et al., 2015, Li and Dong, 2013), with achievements like AlphaGo, the first DNN program to beat in March 2016 a 9-dan Go Master (Gibney, 2016). Well-tested DNN software libraries are also becoming widely available (like Google's TensorFlow library (Torres, 2016)), as well as highly-specialized, highly-parallelized hardware (like the 16-bits arithmetic Graphical Processing Units Pascal from NVIDIA). According to the literature, two possible learning schemes are possible: supervised and unsupervised (see paragraph 1). We focus on the following on supervised Deep Stacking Networks (DSN) and their derivatives, the Kernel-Deep Stacking Networks (K-DSN) and the Tensor-Deep Stacking Networks (T-DSN), which seem to be the most promising architectures of the new wave of ANNs (Hutchinson et al., 2013).

In order to reduce the dimensionality of ANNs, DSN are based on the "stacking" (so their name) of elementary shallow ANNs, called bricks. At each brick corresponds the basic operator M linking input x to output y given by (7):

$$y = M(x) = V \cdot (\sigma(WU))^+ \sigma(Wx)$$

i.e. we plug the output of Eq. (6) as the input of Eq. (4). For each brick a cost function $Q(W)$ can then be defined and minimized by a backpropagation algorithm. A typical DSN architecture is given in Figure (1).

In the K-DSN approach, the $\sigma(WU)$ sigmoid term is replaced by a more general $G(U)$ term, induced from a so-called kernel function $K$, with $K(A, B) = G^T(A)G(B)$, resulting in (8):

$$y = M(x) = V \, (G(U))^+ \, G(x) = V\big(G(U)^T G(U)\big)^+ G(U)^T G(x)$$

The kernel function K is parameterized w.r.t. a small number of coefficients, even reduced to one coefficient, as for example the kernel function (9):

$$K(A, B) = \exp(-\frac{\|A-B\|^2}{\rho^2}),$$

that replaces the whole set of $W$ weights of formula (8), and so the corresponding layer of neurons, by a "brick-large" $\rho$ coefficient ($\| \, . \, \|$ denotes a distance in the mathematical sense). With such a drastic reduction of the free parameters, the determination of M, for each individual brick, can become a fully convex problem (Deng et al., 2012, Huang et al., 2013).

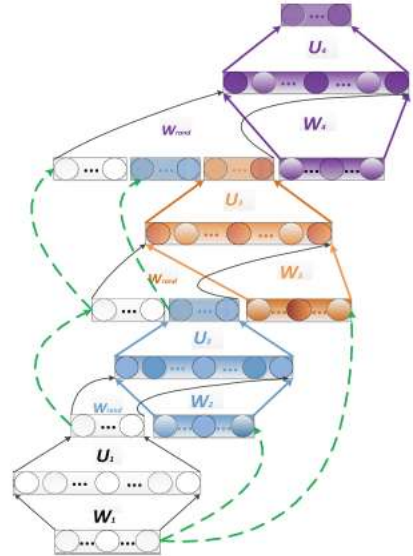

**Figure 1:** A DSN architecture using input–output stacking (Tur et al., 2012). Four basic bricks are illustrated, each with a distinct color. The DSN includes a set of serially connected, overlapping, and layered bricks, wherein each brick has the same architecture defined by Eq. (7). In this figure, the output units of a lower brick are a subset of the input units of an adjacent higher brick in the DSN.

In the T-DSN approach, the set of weights $W$ in Eq. (7) is split in two sub-sets $W_1$ and $W_2$, with Eq. (7) becoming (10):

$$y = M(x) = V \cdot [\sigma(W_1 U) \otimes \sigma(W_2 U)]^+ [\sigma(W_1 x) \otimes \sigma(W_2 x)]$$

where $\otimes$ denotes tensor multiplication. We refer to Hutchinson et al. (2013) for all the technicalities. The authors claim that the main advantage of splitting the weights W into two subsets is the new ability for the T-DSN to take into account higher order covariance statistics in the data (Blanc-Lapierre, 1991). A secondary advantage is that the convexity of the problem improves.

From a theoretical point of view, Equations (8) and (10) can be combined as (11):

$$y = M(x) = V \cdot [G_1(U) \otimes G_2(U)]^+ [G_1(x) \otimes G_2(x)],$$

But we found no trace of practical applications of Eq. (11) in the literature.

### 4. An Artificial Intelligence model layout for the Moorea Ecosystem.

As already stated, it is out of question, to study the time evolution of complicated ecosystems, to derive sets of differential equations. The best that can be envisioned is to be able to set-up "first integrals", i.e. quantities that must be conserved during the evolution of the system (like the conservation of mass for Ecopath).

We have at our disposal three main types of data: (1) time series at different places, like weather recordings, stream gauges readings, coral whitening, pearls oyster diameters for a given oyster age, radioactivity records, mean sea level, biomass for several species, population and housing censuses, etc…(2) single-shot measurements, very often made during acquisition campaigns, like soil drillings, (3) maps, like vegetation maps (classifiers), DTMs, drainage areas, etc…

Time series and maps have very often, on their own, a predictive power. For example, rain and stream gauge readings can be correlated through a convolution integral for the same watershed (Barriot et al., 2016). The same is true for maps: for example, the Universal Soil Loss Equation (USLE) used in erosion studies (Ye et al., 2010) reads (pointwise) $A = R * K * LS * C * P$, where $A$ is the soil loss map, $R$ is the rainfall-runoff erosivity factor map, $K$ is a soil erodibility factor map, $L$ is a slope-length factor map (from DTM), $S$ is the slope steepness factor (also from DTM), $C$ is a cover-management factor and $P$ is a supporting practice factor (from agriculture / soil management). It should be noted that most of the data sets available to monitor ecological systems are in the form of on-site time series (like pluviometer recordings). We refer to Azencott and Dacunha-Castelle (1986) for the processing of irregular time series. Time series of maps (like the evolution of vegetation cover) are very often quite rare (the cost of an airborne campaign is several orders of magnitude larger than the cost of a pluviometer!). From an operational point-of-view, the best use of maps is probably to be introduced as *information context,* i.e. "almost-constant parameters". The most delicate point is certainly to take into account single-shot measurements. We do not have at this moment the right answer. May be they can also be used as information context.

Our main idea, to generalize the differential set of Equations (1), *and to build our realization of the ecosystem avatar*, is to use a DSN or one of its variants (Eq. (11) is our favorite) as a predictor in time, with the same approach as in paragraph 1, with the context information as part of the inputs, but not of the outputs, in the following supervised training form (12):

$$U = (T(t_i), C)$$
$$V = (T(t_{i+1}))$$

with time $t_i$, $i = 1, .., N$ and T must be understood in the vector sense, gathering all the observation types (rain, census, biomass…) we want to take into account. More specifically, for each brick $(k)$, except the first one, we propose the following architecture, inspired from Huang et al. (2013):

$$y_k = M_k(x_k)$$

where $M_k$ is the operator corresponding to the brick (k) and

$$x_k = (T(t_i), C, y_{k-1})$$

In other words, for the training or prediction, the input of each brick, except the first one, is made by concatenating, in the vector sense, the replicated input of the first brick, the context information and the output of the previous brick. For example, if we have 40 different time series, we have 40 corresponding entries for each brick (one for each time series at time $t_i$). The notation $C$ represents the context maps, or context information, with the following rule: for each map, one pixel value = one brick entry. For example, if we have a 100 m resolution DTM over a 10 km² area, this means adding 10,000 entries for each brick. But each brick has only 40 output values, one for each time series. This corresponds to 10,080 entries in the vector $x_k$. It is clear that context maps account for most of the entries in each brick. This is why K-DSN have an edge here, as they only require the computation of "distances" between input data, and not the non-convex determination of a full set of weights per brick (see Eq. 7). From a practical point of view, we will have to convert all inputs $x_k$ to adimensional numbers, with a dimensional scaling factor $\rho$ (Eq. 9) for each data set (time series or context information). For example, if we have a K-DSN with four bricks, we will have for the previous example 41 scaling factors (one for each time series and one for the DTM). We can have scaling factors that vary for each brick, leading to 328 scaling factors to be solved. If we take into account that we also have to regularize the determination of the pseudo-inverse for each brick, this induces 4 additional $\lambda$ ridge regression coefficients (Eq. 5). If now we imagine that we have 10 years of time series with one data point per day, we end up with 10 x 40 x 365 = 146,000 data points plus the 10,000 pixel numbers for the DTM. This means 156,000 data points at large for 332 unknowns to be solved (for Eq. 11). The optimization problem is clearly overdetermined and numerically well-posed thanks to the pseudo-inverse regularization. Additional conservation equations (like the mass conservation of Ecopath, see paragraph 1) will also improve the well-posedness.

## 5. Stability concerns for the Ecosystem Avatar

If we want to use the K-DSN as an integrator over time (in the differential equations meaning of paragraph 1), we have to take into account the stability of the integrator, i.e. the boundedness and accuracy of the successive iterations:

$$T(t_{i+1}) = M(t_i) = M\left(M\left(t_{i-1}\right)\right) = M\left(M\left(M\left(T(t_{i-2})\right)\right)\right) = \cdots.$$

Where M is acting on the time series $T$. When M is linear (Eq. (4)), this stability depends on the spectral radius of M. In the non-linear case (e.g. Eq. 11), we can numerically infer the boundedness and stability by splitting the observed time series in two successive parts: the first one will be used as a training set, and the consecutive part as "ground-true" for deriving by hand a time-horizon beyond which the prediction is "unreliable".

## 6. Conclusion

We proposed in this paper a possible architecture for a simplified version of an *ecosystem avatar*, defined to be a *predictor w.r.t time* (i.e. able to predict the future evolution of the ecosystem), based on a *set of time series* summarizing our knowledge of the past of the ecosystem, and *context information* providing global information about the environment of the ecosystem. This architecture has still to be "ground" tested. This implies to collect and validate all the available data relative to the Moorea island, to validate them, and to test as described in paragraph 5 all the possible avatar architectures. As stated by Bengio (2016): "The science of machine learning is largely experimental because no universal algorithm exists […]. Any knowledge-acquisition algorithm needs to be tested [..] on data specific to the situation at hand […]. There is no other way to prove that it will be consistently better across the board for any given situation than all other algorithms."